\PassOptionsToPackage{twoside=false}{geometry}
\documentclass[pdflatex,sn-nature]{sn-jnl}

\usepackage{amsmath,amssymb,amsfonts}%
\usepackage{amsthm}%
\usepackage{booktabs}
\usepackage{subcaption}
\usepackage{multirow}

\usepackage{graphicx} 

\begin{document}

\title[ASDFormer: A Transformer with Mixtures of Pooling-Classifier Experts for Robust Autism Diagnosis and Biomarker Discovery]{ASDFormer: A Transformer with Mixtures of Pooling-Classifier Experts for Robust Autism Diagnosis and Biomarker Discovery}

\author[1]{\fnm{Mohammad} \sur{Izadi}}\email{m.izadi@alumni.iut.ac.ir}

\author*[1]{\fnm{Mehran} \sur{Safayani}}\email{safayani@iut.ac.ir}

\affil[1]{\orgdiv{Department of Electrical and Computer Engineering}, \orgname{Isfahan University of Technology}, \orgaddress{\city{Isfahan} \postcode{84156-83111}, \country{Iran}}}

\abstract{	
Autism Spectrum Disorder (ASD) is a complex neurodevelopmental condition marked by disruptions in brain connectivity. Functional MRI (fMRI) offers a non-invasive window into large-scale neural dynamics by measuring blood-oxygen-level-dependent (BOLD) signals across the brain. These signals can be modeled as interactions among Regions of Interest (ROIs), which are grouped into functional communities based on their underlying roles in brain function. Emerging evidence suggests that connectivity patterns within and between these communities are particularly sensitive to ASD-related alterations. Effectively capturing these patterns and identifying interactions that deviate from typical development is essential for improving ASD diagnosis and enabling biomarker discovery. In this work, we introduce ASDFormer, a Transformer-based architecture that incorporates a Mixture of Pooling-Classifier Experts (MoE) to capture neural signatures associated with ASD. By integrating multiple specialized expert branches with attention mechanisms, ASDFormer adaptively emphasizes different brain regions and connectivity patterns relevant to autism. This enables both improved classification performance and more interpretable identification of disorder-related biomarkers. Applied to the ABIDE dataset, ASDFormer achieves state-of-the-art diagnostic accuracy and reveals robust insights into functional connectivity disruptions linked to ASD, highlighting its potential as a tool for biomarker discovery.
}


\maketitle

\section{Introduction}\label{sec1}

Autism Spectrum Disorder (ASD) is a neurodevelopmental condition marked by deficits in social interaction, communication, and restrictive behaviors, typically emerging in early childhood~\cite{first,second}. Early diagnosis is crucial for effective intervention, but conventional diagnostic practices rely on subjective behavioral assessments, which are time-consuming and prone to variability~\cite{third,Gao2024}. 

Neuroimaging, particularly resting-state functional magnetic resonance imaging (rs-fMRI), provides an objective alternative to traditional diagnostic methods by capturing spontaneous BOLD fluctuations without task requirements, making it ideal for pediatric and neurodiverse populations \cite{FNSurvey2024,fmriDecodingSurvey2022}. In rs-fMRI analysis, the brain is often parcellated into regions of interest (ROIs), which represent functionally distinct areas. Functional connectivity is then assessed by calculating pairwise correlations, typically using Pearson correlation, between the time series of each ROI. This method quantifies the temporal relationships between different brain regions, offering insights into the organization and disruption of brain networks in ASD~\cite{interpretTotal}. Given the complexity and high dimensionality of rs-fMRI data, machine learning has become essential for ASD classification. Data-driven models leveraging functional connectivity graphs offer a more comprehensive approach for identifying ASD-related neural alterations.

Recent advances in deep learning have greatly enhanced the modeling of brain connectivity patterns in ASD. Graph-based methods, particularly Graph Neural Networks (GNNs), have shown promise in capturing relationships between brain regions~\cite{BrainGNN2021,Park2023}. However, Transformer architectures, with their self-attention mechanisms, offer a more powerful approach by directly modeling interactions across distant brain regions, capturing long-range dependencies and providing a global view of brain network dynamics~\cite{SamplingRef,Kumar2024}. Recent studies highlight their effectiveness in identifying atypical interactions among large-scale networks, such as the default mode and sensorimotor systems, which are consistently implicated in ASD~\cite{Communityaware2023}. 
Transformer architectures offer interpretability through attention mechanisms that highlight prominent interactions among brain regions during prediction. However, attention weights do not guarantee which Regions of Interest (ROIs) are truly responsible for the model’s final decision. As a result, complementary interpretability strategies are necessary to more accurately identify ASD-relevant neural features and support reliable biomarker discovery.

Building on this foundation, Mixture of Experts (MoE) architectures introduce conditional computation via specialized expert subnetworks, enabling scalable learning and improved representational capacity. By dynamically routing inputs to relevant experts, MoEs have traditionally been used to scale the feed-forward networks within Transformer architectures~\cite{MoESurvey}. In brain connectivity modeling, MoEs have the potential to complement interpretable spatialized frameworks, where the focus is on interpretability and spacialization rather than computational efficiency.

In this work, we present \textbf{ASDFormer}, a novel transformer-based framework for functional connectivity modeling in Autism Spectrum Disorder (ASD), with a primary focus on advancing interpretability—a critical yet underexplored aspect in recent transformer-based neuroimaging studies. Our main contributions are:
\begin{itemize}
	\item We propose a Mixture-of-Experts (MoE) decoder that guarantees the identification of specific Regions of Interest (ROIs) directly associated with the model’s final decision. By integrating expert specialization, sparse pooling, and classification with distinct top-$k$ ROI selection criteria, our model accurately captures individualized regional alterations, enabling precise and interpretable ASD diagnosis across heterogeneous subject presentations.
	
	\item We propose a complementary interpretability framework by combining MoE-driven region selection with attention-based transformer modeling. This integration enables both the identification of key diagnostic ROIs and the interpretation of how their representations are influenced by broader connectivity patterns, providing insights into community-level disruptions and candidate biomarkers in ASD.
\end{itemize}

Applied to the ABIDE dataset, our model achieves state-of-the-art performance and reveals connectivity patterns consistent with well-established ASD biomarkers. These results underscore the potential of our specialized framework for advancing ASD diagnosis.

The remainder of this paper is organized as follows. Section 2 reviews related work on ASD classification and mixture-of-experts. Section 3 covers theoretical preliminaries. Section 4 describes our proposed method. Section 5 presents experiments and interpretability analyses on the ABIDE dataset. We conclude with a discussion of implications and future directions.

\section{Related Works}\label{sec2}

\subsection{Autism Spectrum Disorder Modeling}

Graph Neural Networks (GNNs) have been extensively utilized for modeling brain connectivity from fMRI data. ROI-aware methods, such as BrainGNN \cite{BrainGNN2021}, enhance interpretability by learning region-specific representations and incorporating region pooling. Notably, Park and Cho \cite{Park2023} demonstrated that GNNs effectively capture interactions between the superior temporal sulcus and visual cortex, resulting in improved ASD classification.

However, GNNs primarily rely on local message passing, which restricts their ability to model long-range dependencies across the brain network. Transformers overcome this limitation by representing ROIs as input tokens and employing self-attention mechanisms to capture global interactions. This global modeling capacity facilitates more accurate biomarker discovery and ASD diagnosis.

Recent studies have validated the promise of transformers for connectome analysis. Deng et al.~\cite{STtransformer2022} introduced a spatial-temporal attention model (LSTMA) combined with data balancing techniques to enhance ASD prediction. Similarly, Com-BrainTF~\cite{Communityaware2023} integrates local and global attention through a community-aware design, improving both interpretability and hierarchical brain modeling. Although transformers address the local message passing constraints of GNNs and generally achieve superior accuracy, attention maps alone do not guarantee interpretability, since high attention weights may not correspond to the discriminative ROIs that drive classification decisions or reflect ASD-related neural alterations.

\subsection{Mixture of Experts}
Mixture-of-Experts (MoE) architectures offer a scalable framework for increasing model capacity via conditional computation \cite{huang2024toward}. Shazeer et al.~\cite{MoEloss} introduced the sparsely-gated MoE, where a learned gating mechanism activates a small subset of expert networks per input. This enables significant expansion in model size with modest computational overhead, though it requires auxiliary regularization to ensure balanced expert utilization.

Zhou et al.~\cite{MoEECR} proposed \emph{expert choice routing}, in which experts select the top-k tokens instead of tokens selecting experts. This design ensures uniform expert load and accelerates training, achieving strong performance across downstream benchmarks. However, token-level sparsity may leave some inputs underrepresented during training.

Dai et al.~\cite{deepSeek} introduced \emph{DeepSeekMoE}, an MoE architecture explicitly designed for maximal expert specialization. It combines \emph{fine-grained expert segmentation}, which partitions experts into smaller, more specialized units to enable precise and flexible routing, with \emph{shared expert isolation}, which reserves a subset of always-active experts to consolidate common knowledge and eliminate redundancy. This design substantially improves parameter efficiency and knowledge disentanglement, allowing routed experts to focus on distinct, non-overlapping competencies. Scaled from 2B to 145B parameters, DeepSeekMoE matches or surpasses dense model performance while using a fraction of the computational budget, consistently outperforming prior MoE variants such as GShard.

MoE has also demonstrated utility beyond language modeling. Cao et al.~\cite{MoEModal} applied a gated MoE framework in multi-modal image fusion, where local and global experts specialize in distinct modalities (infrared and visible), achieving adaptive, sample-specific integration. This illustrates MoE’s potential in structured or spatially partitioned domains, where full input visibility by each expert is unnecessary.

Building on this foundation, we propose a novel Mixture-of-Experts (MoE)–based decoder, in which each expert performs a distinct sparse pooling operation followed by a dedicated output MLP. This design enables diverse, specialized pathways for aggregating and classifying information from the complete brain representation, thereby offering new interpretive insights into ASD-related alterations across brain regions.

\section{Preliminaries}\label{sec3}

\subsection{Transformer Architecture}
\label{sec:prelimTrans}

Transformers have become the backbone of modern neural architectures for structured data modeling, due to their ability to capture global dependencies through self-attention mechanisms. Given an input $x \in \mathbb{R}^{B \times n \times d}$, where $B$ is the batch size, $n$ the number of tokens (e.g., ROIs), and $d$ the input feature dimension, the transformer encoder produces an output representation $H \in \mathbb{R}^{B \times n \times d}$.

Each self-attention layer operates as follows:
\begin{equation}
	Q = XW_Q, \quad K = XW_K, \quad V = XW_V,
\end{equation}
\begin{equation}
	A = \frac{QK^\top}{\sqrt{d'}}, \quad \text{Attn}(X) = \text{softmax}(A)V,
\end{equation}
where $W_Q, W_K, W_V \in \mathbb{R}^{d \times d'}$ are learned projection matrices, and $d'$ is the attention head dimension. This operation allows each token to attend to all others, enabling global context modeling.

To further increase expressiveness, the transformer uses \emph{multi-head attention} (MHA), which computes $h$ parallel attention heads:
\begin{equation}
	\text{MHA}(X) = \text{Concat}(\text{head}_1, \dots, \text{head}_h)W^O,
\end{equation}
where each head corresponds to separate projections, and $W^O \in \mathbb{R}^{hd' \times d}$ is a learned output projection.

Each attention sublayer is followed by a residual connection and layer normalization:
\begin{equation}
	X' = \text{LayerNorm}(X + \text{MHA}(X)).
\end{equation}

\begin{figure*}[!t]
	\centering
	\begin{subfigure}[b]{0.43\textwidth}
		\includegraphics[height=10cm]{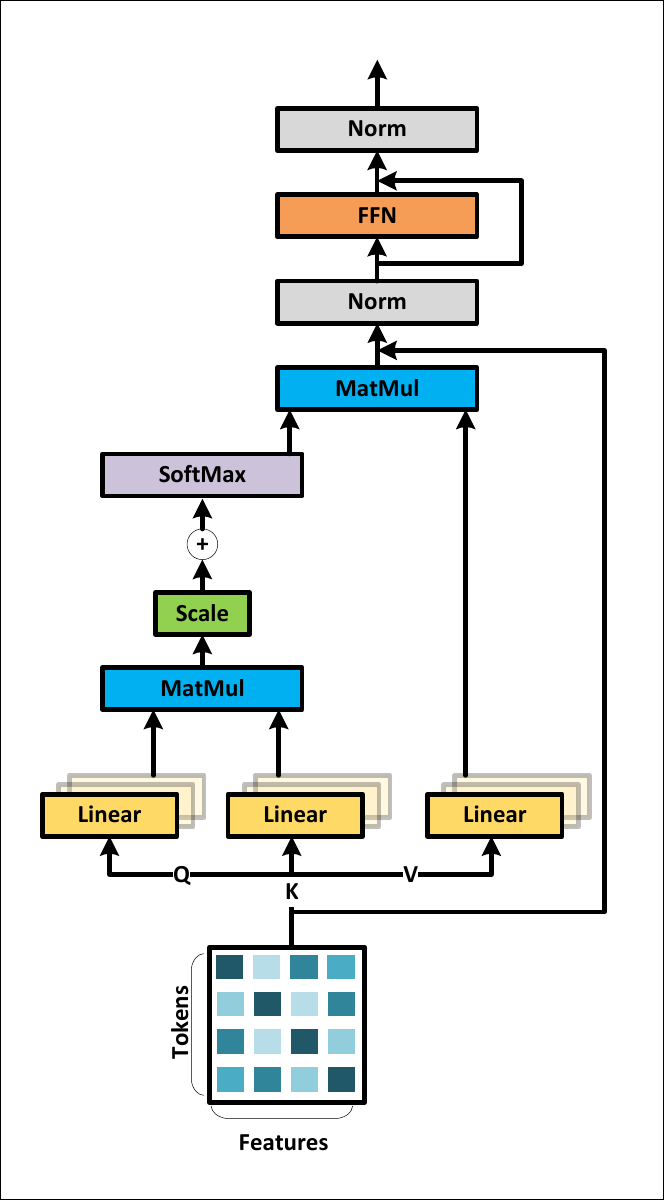}
	\end{subfigure}
	\hfill
	\begin{subfigure}[b]{0.55\textwidth}
		\includegraphics[height=10cm, width=\linewidth]{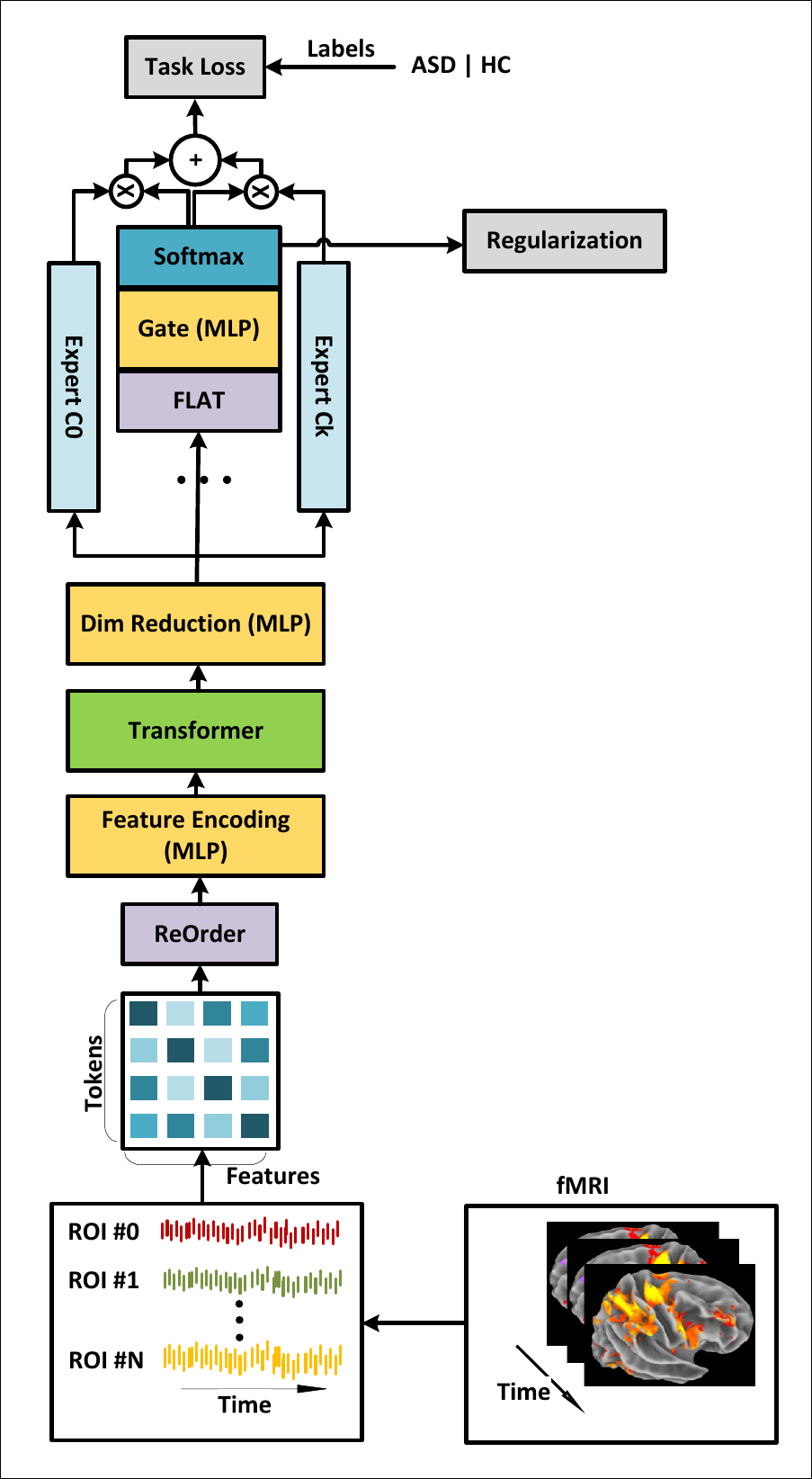}
	\end{subfigure}
	
	\caption{(a) Standard transformer exhibits dense, global attention. 
		(b) Our model introduces modular, sparse specialization via expert-driven selection and decoding.}
	\label{fig:Model}
\end{figure*}

This is followed by a position-wise feed-forward network (FFN) applied independently to each token:
\begin{equation}
	\text{FFN}(X') = \text{GELU}(X'W_1 + b_1)W_2 + b_2,
\end{equation}
\begin{equation}
	H = \text{LayerNorm}(X' + \text{FFN}(X')).
\end{equation}
The complete encoder layer, consisting of MHA, FFN, residual connections, and normalization, forms the backbone of the transformer architecture and is illustrated in Figure~\ref{fig:Model}a. This structure allows flexible, hierarchical modeling of interactions across all ROIs, and serves as a foundation for our proposed extensions.

\section{Proposed Method}\label{sec4}
\label{sec:method}
We propose \textbf{ASDFormer}, a novel transformer-based framework for modeling resting-state functional connectomes, designed to (i) accurately classify individuals with Autism Spectrum Disorder (ASD) versus healthy controls (HC), and (ii) uncover discriminative connectivity patterns and discover key brain regions of interest (ROIs) that are indicative of the final classification decision, serving as candidate neurobiomarkers. The full architecture is depicted in Figure~\ref{fig:Model}b.

\subsection{Functional Connectome Representation}

For each subject, resting-state fMRI (rs-fMRI) data is preprocessed and parcellated into \( N \) regions of interest (ROIs) using a standardized brain atlas. We extract the BOLD time series from each ROI and compute pairwise Pearson correlation coefficients across all ROI pairs, forming a symmetric functional connectivity (FC) matrix of size \( N \times N \) per subject. Each element encodes the degree of temporal synchronization between two regions.
The dataset is represented as a 3D tensor \( \mathbf{X} \in \mathbb{R}^{B \times N \times N} \), where \( B \) is the batch size and \( N \) is the number of brain regions (ROIs). Each subject’s functional connectivity (FC) matrix \( \mathbf{X}_b \in \mathbb{R}^{N \times N} \) captures pairwise correlations between ROIs. We treat each row (or column) of \( \mathbf{X}_b \) as a token, corresponding to the connectivity profile of a specific ROI with all other regions. This formulation yields \( N \) tokens per subject, each represented by an \( N \)-dimensional feature vector.

\subsection{Feature Embedding and Transformer Encoder}

To enable effective learning of high-level representations, we first project the raw ROI features into a latent embedding space via a shared multi-layer perceptron (MLP). This step enhances the model's capacity to extract discriminative features by introducing non-linear transformations. Concretely, for the \( i^\text{th} \) ROI of the \( b^\text{th} \) subject, the embedded feature vector is computed as:

\begin{equation}
	\mathbf{z}_{b,i} = \text{LayerNorm}\left(\text{MLP}(\mathbf{x}_{b,i})\right), \quad \mathbf{z}_{b,i} \in \mathbb{R}^{d},
\end{equation}

where \( \mathbf{x}_{b,i} \in \mathbb{R}^{N} \) is the original connectivity-based feature vector and \( d \) is the embedding dimension. The subsequent application of layer normalization stabilizes the input distribution to the transformer and improves training dynamics. This operation yields an embedded token sequence \( \mathbf{Z} \in \mathbb{R}^{B \times N \times d} \).

Next, we employ a transformer encoder to model contextual dependencies across ROIs. The self-attention mechanism allows each ROI to aggregate information from all others, facilitating the modeling of global brain-wide interactions that are often disrupted in ASD. The transformer produces a set of contextualized ROI representations:

\begin{equation}
	\mathbf{H} = \text{Transformer}(\mathbf{Z}) \in \mathbb{R}^{B \times N \times d},
\end{equation}

where \( \mathbf{h}_{b,i} \in \mathbb{R}^{d} \) denotes the contextualized embedding for the \( i^\text{th} \) ROI of subject \( b \). These representations encode both local connectivity features and their global inter-regional dependencies. The resulting contextual embeddings are subsequently passed to a decoder module responsible for performing subject-level classification and identifying brain regions critical to the model’s decision-making process.

\subsection{Mixture-of-Experts Decoder with Sparse Attention Pooling}
\label{sec:moe}

To improve interpretability and enable expert specialization, we introduce a Mixture-of-Experts (MoE) decoder with sparse attention pooling. This module dynamically selects and combines information from multiple experts.

\paragraph{Input Representation.}
Let \( \mathbf{H} \in \mathbb{R}^{B \times N \times d} \) denote the output of the transformer encoder, where \( \mathbf{H}_b \in \mathbb{R}^{N \times d} \) represents the set of contextualized embeddings for all ROIs of the \( b^\text{th} \) subject. To reduce computational overhead, particularly for the gating network, we apply a shared dimensionality reduction MLP across all ROIs:
\begin{equation}
	\mathbf{H}' = \text{MLP}_{\text{red}}(\mathbf{H}) \in \mathbb{R}^{B \times N \times d'}, \quad \text{with } d' < d.
\end{equation}
Each ROI embedding \( \mathbf{h}'_{b,i} \in \mathbb{R}^{d'} \) now represents a compact summary of the contextualized transformer output for ROI \( i \) in subject \( b \).

\paragraph{Sparse Attention Pooling Experts.}
We define \( E \) experts, each using an attention-based top-\( k \) pooling scheme to identify salient ROIs per subject. For each expert \( e \in \{1, \dots, E\} \), a lightweight MLP \( f_{\text{attn}}^{(e)}: \mathbb{R}^{d'} \to \mathbb{R} \) maps each ROI embedding to a scalar attention logit:
\begin{equation}
	\alpha_{b,i}^{(e)} = f_{\text{attn}}^{(e)}(\mathbf{h}'_{b,i}), \quad \text{for } i = 1, \dots, N.
\end{equation}

For each subject \( b \), the top-\( k \) ROI indices selected by expert \( e \) are defined as:
\begin{equation}
	\mathcal{K}_b^{(e)} = \text{TopK}(\{\alpha_{b,i}^{(e)}\}_{i=1}^{N}, k),
\end{equation}
where \( \text{TopK}(\cdot, k) \) returns the indices of the \( k \) largest elements.

We then apply a masked softmax over these selected logits to obtain normalized attention weights:
\begin{equation}
	w_{b,i}^{(e)} = \frac{\exp(\alpha_{b,i}^{(e)})}{\sum_{j \in \mathcal{K}_b^{(e)}} \exp(\alpha_{b,j}^{(e)})} \cdot \mathbb{I}[i \in \mathcal{K}_b^{(e)}],
\end{equation}
where \( \mathbb{I}[\cdot] \) is the indicator function, equal to 1 for the \(k\) selected ROIs and 0 otherwise. These weights are used to aggregate the top-\( k \) ROI embeddings into a single pooled representation:
\begin{equation}
	\mathbf{z}_b^{(e)} = \sum_{i=1}^{N} w_{b,i}^{(e)} \cdot \mathbf{h}'_{b,i} \in \mathbb{R}^{d'}.
\end{equation}

Each expert contains a dedicated classifier MLP, denoted \( f_{\text{cls}}^{(e)} \), which maps the pooled embedding \( \mathbf{z}_b^{(e)} \) to output logits for the classification task. Specifically:
\begin{equation}
	\mathbf{y}_b^{(e)} = f_{\text{cls}}^{(e)}(\mathbf{z}_b^{(e)}) \in \mathbb{R}^{C},
\end{equation}
where \( C \) is the number of output classes. In the case of ASD diagnosis, we set \( C = 2 \), corresponding to the categories \textit{ASD} and \textit{HC} (healthy control).

\paragraph{Gating Network.}
To combine expert outputs in a data-dependent manner, we employ a shared gating network. For each subject \( b \), the reduced embeddings \( \mathbf{H}'_b \in \mathbb{R}^{N \times d'} \) are flattened:
\begin{equation}
	\mathbf{v}_b = \text{vec}(\mathbf{H}'_b) \in \mathbb{R}^{N \cdot d'}.
\end{equation}
The gating MLP maps \( \mathbf{v}_b \) to expert selection logits:
\begin{equation}
	\mathbf{g}_b = f_{\text{gate}}(\mathbf{v}_b) \in \mathbb{R}^{E},
\end{equation}
and computes normalized gating weights via softmax:
\begin{equation}
	\pi_b^{(e)} = \frac{\exp(g_b^{(e)})}{\sum_{j=1}^E \exp(g_b^{(j)})}, \quad \text{for } e = 1, \dots, E.
\end{equation}
The final prediction for each subject is a weighted sum of expert logits:
\begin{equation}
	\mathbf{y}_b^{\text{final}} = \sum_{e=1}^{E} \pi_b^{(e)} \cdot \mathbf{y}_b^{(e)} \in \mathbb{R}^{C}.
\end{equation}

\subsection{Loss Function}

Our training objective combines the standard cross-entropy classification loss with a regularization term designed to prevent expert collapse in the Mixture-of-Experts (MoE) framework. Although the use of sparse top-$k$ pooling naturally encourages expert specialization, some experts may receive disproportionately low attention weights across the training batch, limiting their contribution and model robustness.

To address this, we incorporate a \emph{regularization loss} that promotes balanced expert utilization by penalizing high variance in expert importance, following the approach in \cite{MoEloss}.
Formally, let $\mathbf{g} \in \mathbb{R}^{B \times E}$ denote the gating weights across a batch of size $B$ for $E$ experts, where $g_{b,e}$ represents the attention weight of expert $e$ for sample $b$. The importance of each expert is computed as:
\begin{equation}
	I_e = \sum_{b=1}^B g_{b,e}
\end{equation}
We then compute the coefficient of variation squared ($\mathrm{CV}^2$) of expert importance:
\begin{equation}
	\mathrm{CV}^2 = \left(\frac{\sigma(I)}{\mu(I) + \varepsilon}\right)^2
\end{equation}
where $\mu(I)$ and $\sigma(I)$ are the mean and standard deviation of the importance vector $\mathbf{I} = [I_1, \ldots, I_E]$, and $\varepsilon$ is a small constant for numerical stability.

The final loss is defined as:
\begin{equation}
	\mathcal{L} = \mathcal{L}_{\mathrm{CE}} + \lambda \, \mathrm{CV}^2
\end{equation}
where $\mathcal{L}_{\mathrm{CE}}$ is the cross-entropy loss for classification, and $\lambda$ is a regularization coefficient controlling the strength of the collapse penalty.

This combined loss encourages the model to avoid consistently selecting a single expert or a small subset of experts.

\section{Experiments}\label{sec5}

\subsection{Datasets and Experimental Settings}

\textbf{Datasets:} We use the ABIDE dataset, comprising 1009 resting-state fMRI scans (51.14\% ASD) from 17 international sites~\cite{ABIDE}. Preprocessing followed the CPAC pipeline with band-pass filtering (0.01–0.1 Hz) and no global signal regression. Brain parcellation was performed using the Craddock 200 atlas~\cite{Craddock200}, yielding 200 ROIs grouped into eight functional communities~\cite{NetworkRef}: cerebellum and subcortical structures (CS \& SB), visual network (V), somatomotor network (SMN), dorsal attention network (DAN), ventral attention network (VAN), limbic network (L), frontoparietal network (FPN), and default mode network (DMN). Functional connectivity (FC) matrices were computed via Pearson correlation of mean ROI time series. The dataset was split into training (70\%), validation (10\%), and test (20\%) sets using stratified sampling~\cite{SamplingRef}.

\textbf{Experimental Settings:} Models were implemented in PyTorch and trained on a single NVIDIA RTX 2060 Super GPU. The architecture used 8 attention heads and a 200-dimensional embedding. Optimization was done via Adam with a learning rate and weight decay of \(10^{-4}\), batch size 64, and early stopping based on validation AUROC, over a maximum of 50 epochs. All MLPs in the network consist of two layers with GELU activation and component-specific hidden dimensions. We framed ASD vs. HC classification as a binary task and evaluated performance using AUROC, accuracy, sensitivity, and specificity. For the baseline comparison, we report the results across five independent runs. For evaluation and interpretation, we use the model that achieved the highest AUC.

\subsection{Baseline Comparisons}
To rigorously evaluate the effectiveness of our proposed method, we conduct experiments on the ABIDE dataset and compare it against three categories of strong baseline models: 
(i) \textbf{Transformer-based architectures}, including Com-BrainTF~\cite{Communityaware2023} and BrainNetTF~\cite{SamplingRef}; 
(ii) \textbf{Convolutional neural networks on static functional connectivity graphs}, encompassing BrainNetCNN~\cite{BrainNetCNN} and the CNN-based architecture of Firouzi and Fadaei~\cite{CNNCompare}; and 
(iii) \textbf{Neural networks leveraging dynamically inferred brain networks}, exemplified by FBNETGEN~\cite{Fbnetgen}. 

\cite{Fbnetgen, BrainNetCNN, SamplingRef} were selected as state-of-the-art techniques, widely recognized in the literature~\cite{Communityaware2023} for their strong performance on the ABIDE dataset. This selection ensures a comprehensive and rigorous comparative analysis across diverse modeling paradigms.

As presented in Table~\ref{table:results}, our method consistently surpasses all baseline models across most evaluation metrics, highlighting its superior capability in autism spectrum disorder classification.

\begin{table}[h!]
	\centering
	\setlength{\tabcolsep}{8pt}
	\begin{tabular}{lcccc}
		\toprule
		Model & AUC (\%) & Accuracy (\%) & Sensitivity (\%) & Specificity (\%) \\
		\midrule
		FBNETGNN\cite{Fbnetgen} & 72.64 $\pm$ 8.91 & 65.60 $\pm$ 9.72 & 62.19 $\pm$ 12.10 & 67.53 $\pm$ 13.86 \\
		BrainNetCNN\cite{BrainNetCNN} & 76.45 $\pm$ 4.54 & 68.90 $\pm$ 5.48 & 66.45 $\pm$ 13.83 & \textbf{70.99 $\pm$ 7.26} \\
		BrainNetTF\cite{SamplingRef} & 77.58 $\pm$ 6.40 & 68.00 $\pm$ 5.05 & 77.95 $\pm$ 13.62 & 58.89 $\pm$ 20.12 \\
		Com-BrainTF\cite{Communityaware2023} & 78.77 $\pm$ 3.89 & 69.60 $\pm$ 4.07 & 74.50 $\pm$ 8.73 & 65.76 $\pm$ 10.16 \\
		CNN-FC~\cite{CNNCompare} & 72.91 $\pm$ 5.66 & 66.99 $\pm$ 7.68 & 71.27 $\pm$ 7.23 & 62.33 $\pm$ 10.16\\
		\textbf{ASDFormer} & \textbf{81.17 $\pm$ 5.00} & \textbf{74.60 $\pm$ 4.83} & \textbf{82.55 $\pm$ 10.19} & 66.09 $\pm$ 4.74 \\
		
		\bottomrule
	\end{tabular}
	\caption{Comparison of models with mean $\pm$ std for each metric (scaled by 100).}
	\label{table:results}
\end{table}

\begin{figure}[!t]
	\centering
	\begin{subfigure}{0.9\textwidth}
		\centering
		\includegraphics[width=\linewidth]{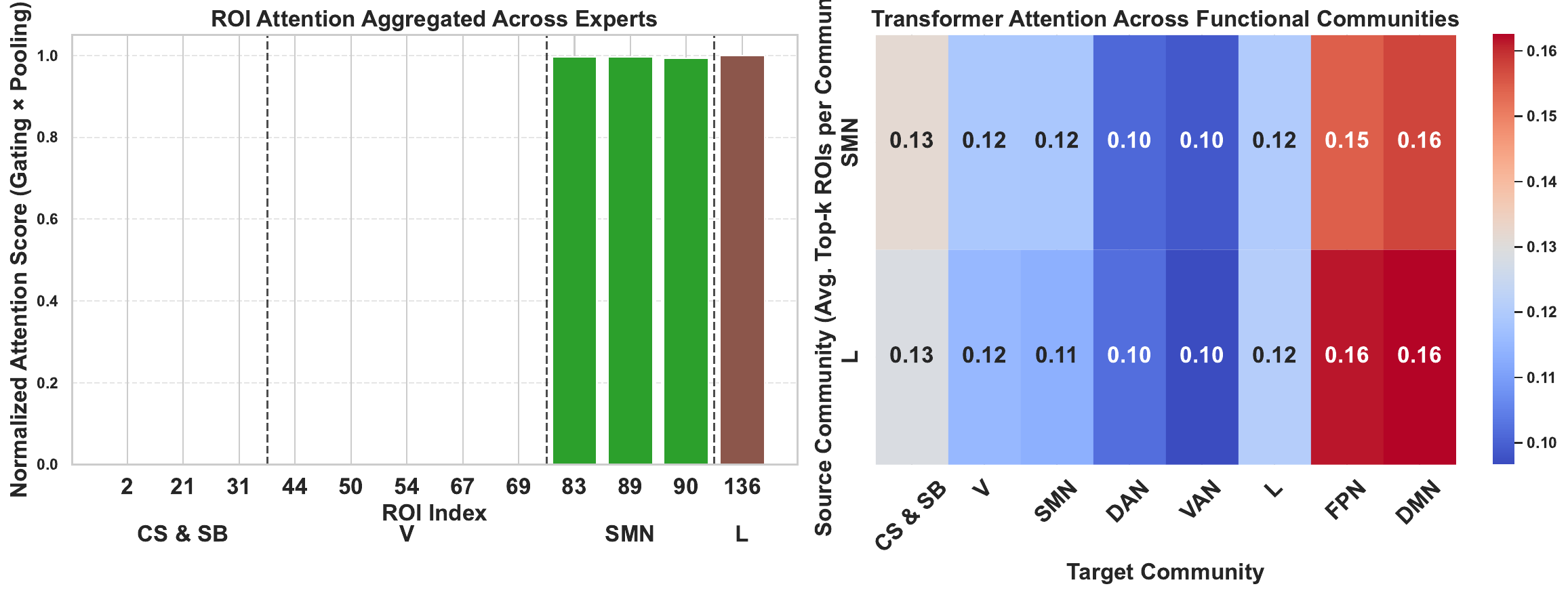}
		\caption{Subject 1}
		\label{fig:subject1}
	\end{subfigure}
	\vspace{0.4cm}
	\begin{subfigure}{0.9\textwidth}
		\centering
		\includegraphics[width=\linewidth]{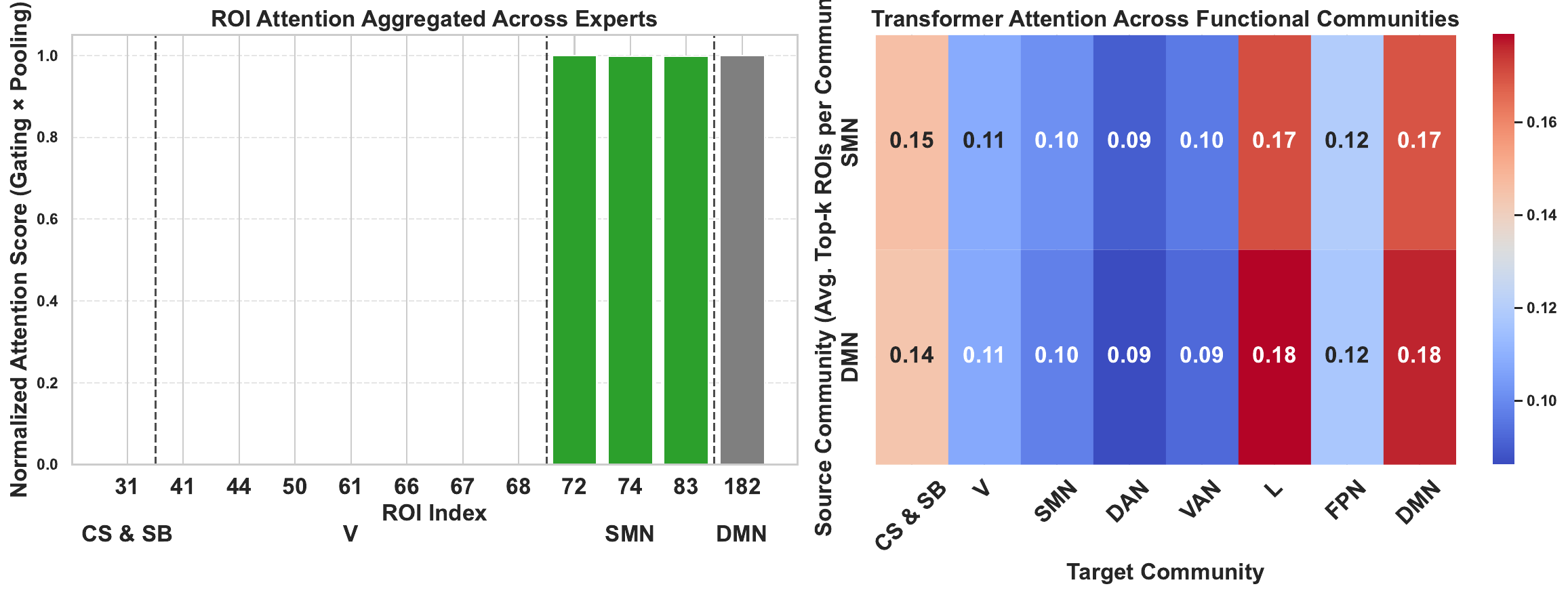}
		\caption{Subject 2}
		\label{fig:subject2}
	\end{subfigure}
	\vspace{0.4cm}
	\begin{subfigure}{0.9\textwidth}
		\centering
		\includegraphics[width=\linewidth]{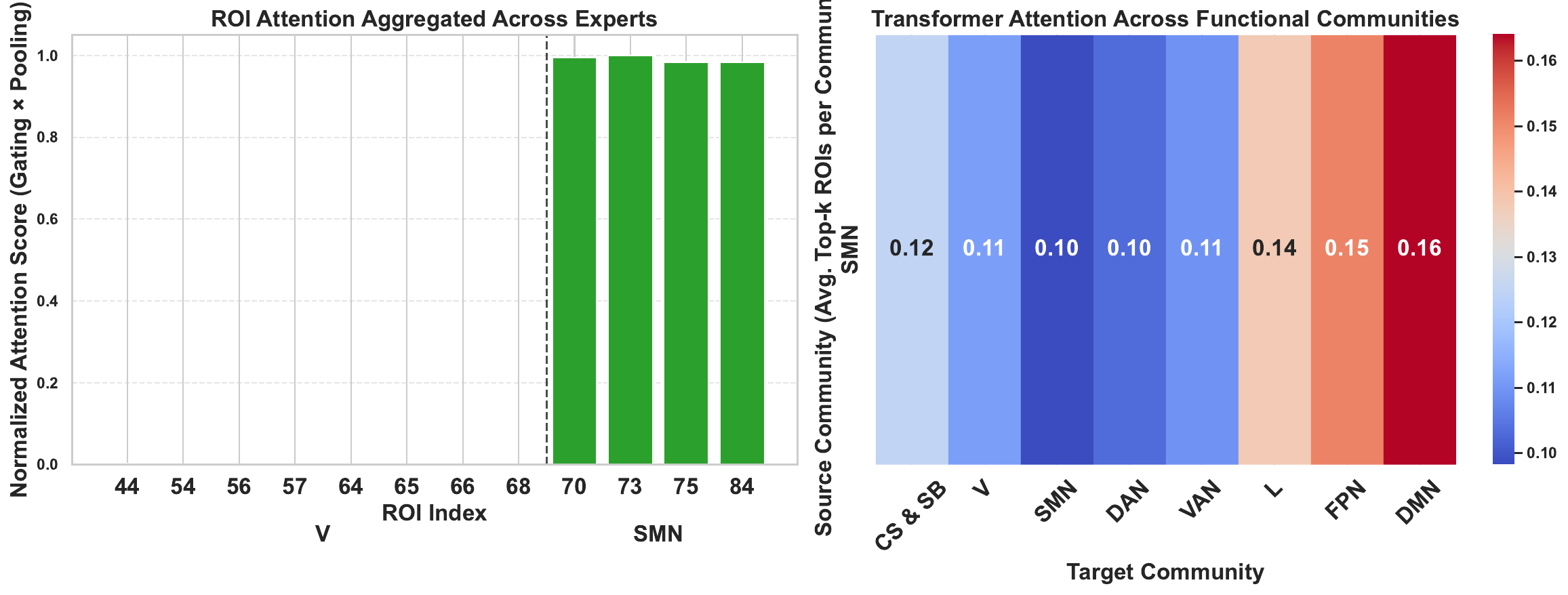}
		\caption{Subject 3}
		\label{fig:subject3}
	\end{subfigure}
	\caption{Interpretability analysis of the ASDFormer model for three distinct ASD subjects. Each subfigure represents a different subject, where the left chart displays the selected tokens as determined by experts, and the right chart visualizes the corresponding attention map produced by the model.}
	\label{fig:interpretation}
\end{figure}

\subsection{Evaluation and Interpretability of ASDFormer}

We perform an analysis of the gate probabilities for both the healthy control (HC) and ASD subjects. This evaluation involves two experts, each with distinct roles in the decision-making process. For the correctly classified HC subject in the test dataset, the gate probabilities after the softmax function are 0.824 for the first expert and 0.176 for the second expert. Conversely, for the correctly classified ASD subject, these probabilities are 0.131 and 0.869, respectively. Notably, the gate weights indicate that the first expert specializes in HC subjects, while the second expert is more effective for ASD detection.

The pooling layer parameter \(k\) is set to 8 for the first expert and 4 for the second, values determined empirically through experimentation. These parameters play a crucial role in expert specialization, with the hidden size and layer count kept constant across experts for simplicity. The number of experts and \(k\) can be further adjusted depending on the dataset and specific requirements. Additionally, the regularization loss coefficient is set to 0.23, ensuring a balance between specialization and generalization.

\paragraph{Biomarker Interpretation:} To further elucidate the interpretability of ASDFormer, we analyze three randomly selected ASD subjects from the correctly classified ASD group and present their biomarker interpretation in Figure~\ref{fig:interpretation}. For each subject, the left chart shows the regions of interest (ROIs) selected by both experts using the top-k decision from their respective pooling layers. The reported scores are computed by multiplying the gate probabilities of each expert with the top-k attention scores corresponding to the relevant expert’s pooling attention.

For ASD subjects, expert 2, with \(k = 4\), exhibits a notably higher gate weight, resulting in significantly more attention being allocated to four specific ROIs, as illustrated in the left charts. The x-axis of these charts corresponds to the ROI indices, allowing us to clearly identify which ROIs are most influential in the model's final decision. Since the transformer layer updates the features of each ROI (token) based on the attention weights assigned to other ROIs, the selection of ROIs in the output layer by each expert can be understood as a result of specific combinations of ROIs and their associated communities. Furthermore, we present the attention relationships between the selected ROIs (from the experts) and other ROIs within the brain network by examining the transformer's attention map (i.e., a row of the attention matrix). This helps highlight the key functional connectivity patterns that contribute to the subject’s classification as ASD. For clarity, in the charts, we average the selected ROIs within each community and compute the average attention received by the target ROIs across the relevant functional communities.

As shown in Figure~\ref{fig:subject1}, ROI indices 83, 89, 90, and 136 exhibit significantly high attention, indicating their influence on the model’s decision. ROIs 83, 89, and 90 belong to the Sensorimotor Network (SMN) community, while ROI 136 is part of the Limbic community. The attention distribution for the SMN ROIs, as illustrated in the right heatmap, reveals that the SMN predominantly focuses on the Fronto-parietal Network (FPN) and Default Mode Network (DMN). Similarly, ROI 136 primarily attends to both FPN and DMN. For the second subject, shown in Figure~\ref{fig:subject2}, ROIs from the SMN and DMN communities exhibit a similar pattern, with SMN ROIs paying the most attention to the Limbic, DMN, and Cerebellum-Sensory and Somatomotor (CS and SB) communities, while DMN ROIs focus on the same communities. For the third subject in Figure~\ref{fig:subject3}, all ROIs belong to the SMN, with attention directed mostly toward DMN, FPN, and Limbic networks. Across these subjects, the SMN, DMN, and Limbic networks contain ROIs critical to the final classification decision.

Prominent associations with ASD were identified in the intra-network connectivity within the Default Mode Network (DMN) (Figure~\ref{fig:subject2}), as well as in inter-network relationships, particularly between the Sensorimotor Network (SMN) and the Fronto-parietal Network (FPN), SMN and DMN, SMN and Limbic, and between DMN and Limbic, DMN and FPN, and SMN and DMN with the Cerebellum-Sensory and Somatomotor (CS \& SB) networks. These results are consistent with prior studies that have emphasized the significance of these connectivity patterns in the context of ASD~\cite{interpretTotal, DMN2017,Aging2024,khan2015cerebro}.

\begin{table}[h!]
	\centering
	\setlength{\tabcolsep}{8pt}
	\begin{tabular}{lcccc}
		\toprule
		Model & AUC (\%) & Accuracy (\%) & Sensitivity (\%) & Specificity (\%) \\
		\midrule
		
		\text{CLS} & 69.84 $\pm$ 5.20 & 64.40 $\pm$ 7.33 & 80.84 $\pm$ 5.29 & 47.89 $\pm$ 15.72 \\
		
		\text{OCRead} & 78.91 $\pm$ 5.96 & 71.20 $\pm$ 6.98 & 79.25 $\pm$ 6.86 & 63.68 $\pm$ 11.42 \\ 
		
		\text{Pooling-Classifier} & 73.07 $\pm$ 7.28 & 65.40 $\pm$ 7.40 & 74.70 $\pm$ 14.99 & 56.75 $\pm$ 13.57 \\

		\textbf{ASDFormer} & \textbf{81.17 $\pm$ 5.00} & \textbf{74.60 $\pm$ 4.83} & \textbf{82.55 $\pm$ 10.19} & \textbf{66.09 $\pm$ 4.74} \\
		
		\bottomrule
	\end{tabular}
	\caption{Ablation study of decoder designs under a fixed single-layer transformer and shared feature encoding MLP. \texttt{CLS} offers interpretability but lowest performance, \texttt{OCREAD} improves accuracy at the cost of interpretability, and \texttt{Pooling-Classifier} preserves interpretability with reduced accuracy. \textbf{ASDFormer} achieves the best performance while maintaining interpretability. Results are mean $\pm$ standard deviation, scaled by 100.}
	
	\label{table:ablation}
\end{table}
\subsection{Ablation Study}

To assess the effectiveness of our proposed decoder, we conducted an ablation study under a controlled setting where the number of transformer layers was fixed to one and all methods employed the same feature encoding MLP. Table~\ref{table:ablation} summarizes the results for the different design choices.

We first adopted the \texttt{CLS} token approach, a common practice in Transformer-based architectures \cite{graphormer2021}, where a learnable token is appended to the input sequence to aggregate information, followed by an MLP classifier. This design (row \textbf{CLS} in Table~\ref{table:ablation}) yielded limited performance (AUC = 69.84\%), despite allowing interpretability through attention visualization—specifically, by inspecting which tokens the \texttt{CLS} token attended to most.

We then evaluated the \texttt{OCRead} strategy, originally introduced in \cite{SamplingRef} and later employed in \cite{Communityaware2023} (row \textbf{OCRead} in Table~\ref{table:ablation}). This method significantly improved performance (AUC = 78.91\%) but sacrificed token-level interpretability, as the final decision could not be directly attributed to specific input tokens.

To preserve interpretability, we implemented the \texttt{Pooling-Classifier} design (row \textbf{Pooling-Classifier} in Table~\ref{table:ablation}), in which attention pooling is followed by an MLP classifier, corresponding to a single-expert configuration without gating. This architecture allows identification of the top-$k$ most influential tokens contributing to the final decision. However, its performance was notably lower (AUC = 73.07\%) compared to alternative designs.

Building upon this, we introduce \textbf{ASDFormer}, which augments the Pooling-Classifier architecture with a Mixture-of-Experts (MoE) module comprising multiple experts and a learnable gating mechanism. To mitigate the risk of the gate consistently selecting a single expert, we introduced a regularization term that promotes balanced expert utilization across the network. In this configuration, a dimensionality-reduction MLP is employed prior to gating, as the gate operates on a flattened representation whose dimensionality would otherwise be prohibitively large. This design retains token-level interpretability while achieving the highest performance among all evaluated configurations (AUC = 81.17\%), thereby meeting both our accuracy and explainability objectives.

\section{Conclusion and Future Works}\label{sec6}

We introduced \textbf{ASDFormer}, a Transformer-based framework enhanced with a Mixture of Experts (MoE) decoder for modeling functional connectivity in Autism Spectrum Disorder (ASD). By combining expert specialization with attention mechanisms, ASDFormer improves both classification accuracy and interpretability. The model identifies critical alterations in functional connectivity, enabling a more individualized and precise analysis of brain network disruptions relevant to ASD diagnosis. When applied to the ABIDE dataset, ASDFormer achieved state-of-the-art diagnostic performance, uncovering connectivity patterns that align with well-established ASD biomarkers. These results highlight the model's potential for early, data-driven diagnosis and its ability to uncover key biomarkers associated with ASD.

Future work will explore enhancements such as the integration of Graphormer \cite{graphormer2021}, community-aware biases, ROI centrality, and direct functional edge biases to further improve the model's performance and interpretability.

\backmatter

\bmhead{Code availability}
The code used in this study is publicly available at: \url{https://github.com/mohammad1997izadi/ASDFormer}.

\bmhead{Declaration of Competing Interest}

The authors declare that they have no known competing financial interests or personal relationships that could have appeared to influence the work reported in this paper.

\bmhead{Acknowledgements}

This research did not receive any specific grant from funding agencies in the public, commercial, or not-for-profit sectors.

\bmhead{Data availability}

Data will be made available on request.

\bibliography{references}

\end{document}